\documentclass{article}

\usepackage[preprint]{colm2026_conference}
\usepackage{fontspec}

\renewcommand{\encodingdefault}{T1}
\normalfont
\usepackage{microtype}
\usepackage{graphicx}
\usepackage{trimclip}
\usepackage{xcolor}
\usepackage{booktabs}
\usepackage{multirow}
\usepackage{array}
\usepackage{colortbl}
\usepackage{wrapfig}
\usepackage{placeins}
\usepackage{float}
\usepackage{amsmath}
\usepackage{amssymb}
\usepackage{marvosym}
\usepackage{tikz}
\usepackage{tcolorbox}
\usepackage{hyperref}
\usepackage{url}

\usepackage{amsmath,amsfonts,bm}

\def\eqref#1{equation~\ref{#1}}

\def\1{\bm{1}}

\DeclareMathAlphabet{\mathsfit}{\encodingdefault}{\sfdefault}{m}{sl}
\SetMathAlphabet{\mathsfit}{bold}{\encodingdefault}{\sfdefault}{bx}{n}

\usepackage{caption}

\newcommand{\gp}[1]{\,\textcolor{red}{\scriptsize($+$#1)}}
\newcommand{\gm}[1]{\,\textcolor{green!45!black}{\scriptsize($-$#1)}}

\definecolor{abyss}{HTML}{121D36}
\definecolor{polarnight}{HTML}{1A2947}
\definecolor{nebula}{HTML}{2B3F66}
\definecolor{steeltrail}{HTML}{6D87BD}
\definecolor{skytrail}{HTML}{8FA8D8}
\definecolor{starlight}{HTML}{DFE7F5}
\definecolor{electricblue}{HTML}{3866FF}
\definecolor{covercream}{HTML}{EEF3FA}
\definecolor{coveraccent}{HTML}{3866FF}

\newfontfamily\outfit[
  Path=./,
  BoldFont=Outfit-SemiBold.ttf
]{Outfit-Regular.ttf}

\hypersetup{
  colorlinks=true,
  linkcolor=electricblue,
  citecolor=electricblue,
  urlcolor=coveraccent,
  filecolor=electricblue
}
\setcitestyle{numbers,square,comma,sort&compress}

\newcommand{\reporttitle}{Learning to \textcolor{electricblue}{Learn from Context}: Synthetic Training from Perturbed Public Documents}
\title{Learning to Learn from Context: Synthetic Training from Perturbed Public Documents}
\author{Haoyi Wu, Yang Xiao, Yusong Sun, Wenyang Hui, Zhaokai Luo, Chengyue Jiang, Mu Chuan}

\begin{document}
\raggedbottom
\fancyhead{}
\renewcommand{\headrulewidth}{0pt}
\color{abyss}
\thispagestyle{empty}

\vspace*{-0.44in}
\begin{tcolorbox}[
  width=\linewidth,
  colback=covercream,
  colframe=covercream,
  boxrule=0pt,
  arc=14pt,
  outer arc=14pt,
  boxsep=0pt,
  left=16pt,
  right=16pt,
  top=9pt,
  bottom=7pt
]
  {\outfit\fontsize{18}{22}\selectfont\bfseries\centering
    \reporttitle\par}
  \vspace{0.9em}
  {\bfseries\centering AllSpark Team\par}

  \vspace{0.5em}
  \begingroup
  \normalfont\small
  \setlength{\parindent}{0pt}
  \setlength{\parskip}{0pt}
Real-world tasks often require large language models (LLMs) to learn from complex task-specific context rather than pretrained parametric knowledge.
This capability remains a weakness of LLMs, while human annotation for such task contexts is expensive and difficult to scale.
Public high-quality documents are an abundant alternative, but much of the public web has already been consumed during pretraining: training on such documents naively would reward memorization rather than context learning.
In this work, we attempt to make use of high-quality public documents with small perturbations and empirically find that LLMs can successfully generate context-dependent reasoning traces and answers, which are then used to train a student model.
Specifically, we construct a synthesis pipeline that (i) rewrites source documents to reduce memorization risk, (ii) generates questions and rubrics that require reasoning over the document, (iii) answers the questions with the document as context, and (iv) admits only samples that genuinely depend on the document.
Without any human annotators, our pipeline generates about 10k samples from 3.5k documents, and the resulting student model substantially improves the performance on CL-bench.
SFT raises a Qwen3.6-35B-A3B student from 13.7\% to 22.8\%, and a subsequent rubric-reward RL stage reaches 24.6\%, on CL-bench comparable with a frontier model of over a trillion parameters, Qwen3.8-2.4T (23.9\%).
We also observe a broad transfer of improvements to long-context understanding, instruction following, and reasoning, while code generation and knowledge remain mostly flat.
We hope this work provides a reproducible and scalable way to improve the ability of LLMs to learn from context, and to facilitate further research on context-grounded reasoning.
  \par
  \endgroup

  \vspace{0.4em}
  \noindent
  \begin{minipage}[b]{0.63\linewidth}
    \outfit\fontsize{8.4}{10.2}\selectfont
    \textbf{Date:} September 27, 2026
  \end{minipage}%
  \hfill
  \begin{minipage}[b]{0.33\linewidth}
    \raggedleft
    \raisebox{-0.30em}{\includegraphics[height=16pt]{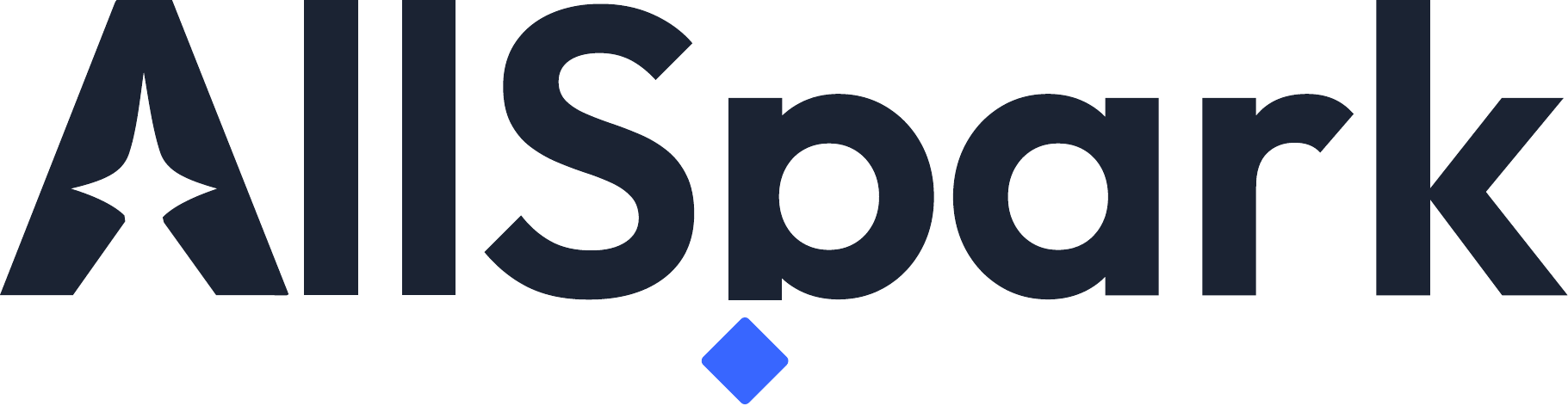}}%
  \end{minipage}
\end{tcolorbox}

\vspace{0.4em}
\begin{figure}[H]
\centering
\includegraphics[width=0.95\linewidth]{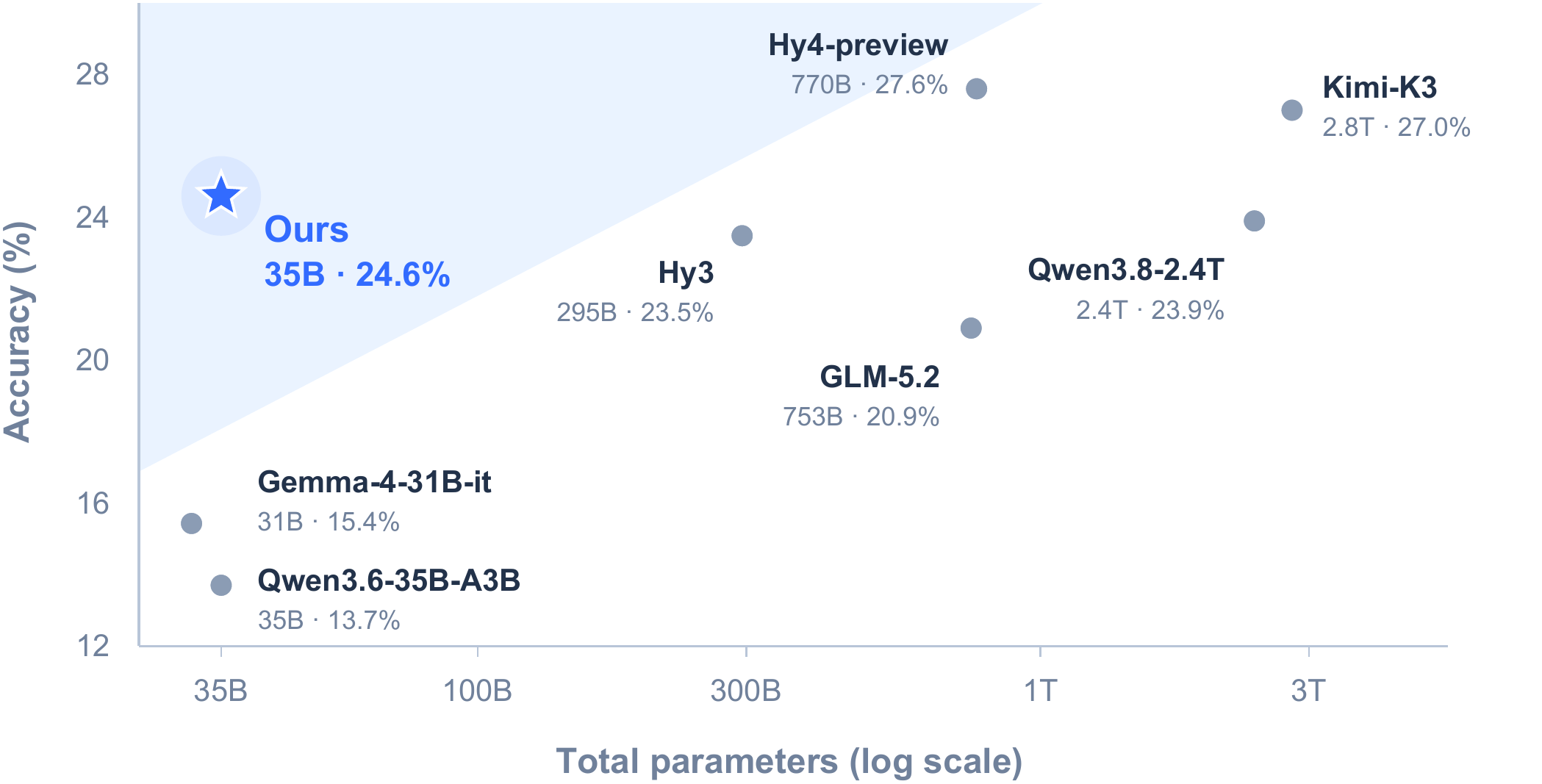}
\caption{CL-bench accuracy versus total model parameters. Our 35B student (blue), trained on 10k synthetic samples from perturbed public documents, reaches 24.6\%, comparable with frontier models that are 10--100$\times$ larger.}
\label{fig:teaser}
\end{figure}

\clearpage

\section{Introduction}
\label{sec:intro}

As the context windows of large language models (LLMs) expand to millions of tokens \citep{zeng2026glm5,qwenteam2026qwen38max,kimiteam2026kimik3}, the ability to learn from context is becoming increasingly important.
LLMs are increasingly deployed in settings where a provided context defines the task: answering questions over internal documents and manuals \citep{zou2024docbench}, reasoning over retrieved passages \citep{lewis2020rag}, interpreting the tool outputs and logs that agents consume \citep{yao2022react,schick2023toolformer}, and checking whether a plan complies with a specification \citep{yao2024taubench}.
This capability, referred to as \emph{context learning} \citep{dou2026clbench}, is distinct from knowledge retrieval: rather than recalling a fact, the model must read a document and reason over it, which requires not only locating information in the text but also understanding the document as a whole and applying it to the task at hand.
Such tasks arise commonly in real-world workflows.

Despite the ubiquity of this setting, context learning remains a weakness of LLMs, particularly for long documents that demand complex reasoning.
Existing long-context benchmarks cover adjacent but distinct capabilities: synthetic retrieval suites such as RULER \citep{hsieh2024ruler} measure effective context windows rather than learning, whereas real-document benchmarks such as LongBench v2 \citep{bai2024longbenchv2} and NoCha \citep{karpinska2024nocha} probe deep understanding but score answers through multiple choice or binary judgments.
Context learning, in contrast, asks whether a model can take up the rules, facts, and procedures that a document defines and apply them faithfully throughout its answer.
CL-bench \citep{dou2026clbench}, a benchmark of 1,899 long-document tasks spanning 18 domains, illustrates the difficulty: even strong frontier models solve fewer than 30\% of the tasks, and the failure mode is characteristic in that models produce answers that appear plausible yet collapse under fine-grained checking.
 
This weakness does not stem from a lack of exposure to long documents, since pretraining corpora contain enough of them to cover most of the domains where context learning matters \citep{gao2020pile,soldaini2024dolma,weber2024redpajama}.
Pretraining teaches models what documents say, but it does not teach them to answer questions \emph{conditioned} on a document, and having read a text therefore does not imply knowing how to use it, not to mention how to apply it to a new task, especially when the document is long and complex.
Instruction-tuning data could in principle provide such supervision, but two obstacles intervene.
First, mainstream instruction datasets are conversational and short-context, and convey little about the faithful use of long documents.
Second, human annotation for long-document tasks is exceptionally expensive: writing a single question requires the annotator with expertise to design a document of tens of thousands of words, verify the answer against it, and specify what a correct answer must contain.

A growing body of work therefore constructs long-context supervision synthetically, yet where the grounding documents come from remains a largely open choice.
One branch grounds supervision in real documents: contexts are either assembled from multiple passages \citep{chen2025longmit,yang2025longfaith,he2025hierarchical}, taken from single long documents in specific domains \citep{lin2025longfinanceqa,pham2025clipper,zhang2024longcite}.
These pipelines, however, may raise memorization concerns, as such publicly available documents are plausible constituents of pretraining corpora \citep{pham2025clipper}.
The other branch synthesizes the documents themselves, for example by constructing tasks over programmatically generated contexts \citep{zhao2024longskywork} or by producing document-grounded and conversational data through prompt-based generation \citep{subramanian2025modular}, which is easy to obtain and scales beyond the length distribution of any curated corpus.
The generated contexts, however, tend to be logically simpler and more homogeneous than real ones, and the gap is measurable: models fine-tuned on synthetic haystacks consistently underperform those fine-tuned on real documents, even though the induced retrieval behavior partially overlaps \citep{zhao2025synthetic}.
Both branches thus have their limitations: real documents carry memorization risk, and synthetic ones are logically simpler and more homogeneous.

In this work, we mitigate the memorization risk of real documents: we perturb them so that the teacher can no longer answer from parametric memory and must instead extract information and reason over the document content.
Specifically, we build a synthesis pipeline on top of small perturbations to public documents.
The pipeline first rewrites each source document with entity renaming and numeric perturbation, then generates questions that require reasoning over the document with LLMs.
It then applies a gap check to admit only samples that genuinely depend on the document, which compares the rubric pass rate of answering with and without the document.
In one memorization audit, a model that answered questions about a well-known technical standard perfectly with no document in context scored nearly zero once the standard was rewritten, because its memorized knowledge collided with the rewritten entities.
Without any human annotators, the pipeline produces about 10k samples from 3.5k public documents, ranging from legal documents to game manuals.

Our experiments show that training on this data yields substantial gains.
Supervised fine-tuning raises a Qwen3.6-35B-A3B student \citep{qwen36_35b_a3b} from 13.7\% to 22.8\% on CL-bench, and a subsequent rubric-reward RL stage reaches 24.6\%, comparable with HY3 \citep{tencent2026hy3} (23.5\%) and Qwen3.8-2.4T \citep{qwenteam2026qwen38max} (23.9\%).
The improvements also transfer beyond the target benchmark: long-context understanding (AALCR \citep{artificialanalysis2025lcr} improves from 62.6 to 69.8), instruction following (IFBench \citep{pyatkin2025generalizing} from 57.7 to 71.7), and reasoning (ARC-AGI-1 \citep{chollet2019arc} from 47.4 to 63.8) all improve substantially, whereas knowledge benchmarks decline slightly (e.g., SimpleQA \citep{wei2024simpleqa} drops from 20.9 to 18.7), consistent with training data that teaches the use of context rather than world knowledge.
We believe this work provides a feasible and scalable path to improving context learning: it turns public documents, even those the model has already read, into effective supervision without any human annotation, and we hope it will facilitate further research on context-grounded reasoning.

\section{The Synthesis Pipeline}
\label{sec:method}

\begin{figure}[t]
\centering
\includegraphics[width=0.95\textwidth]{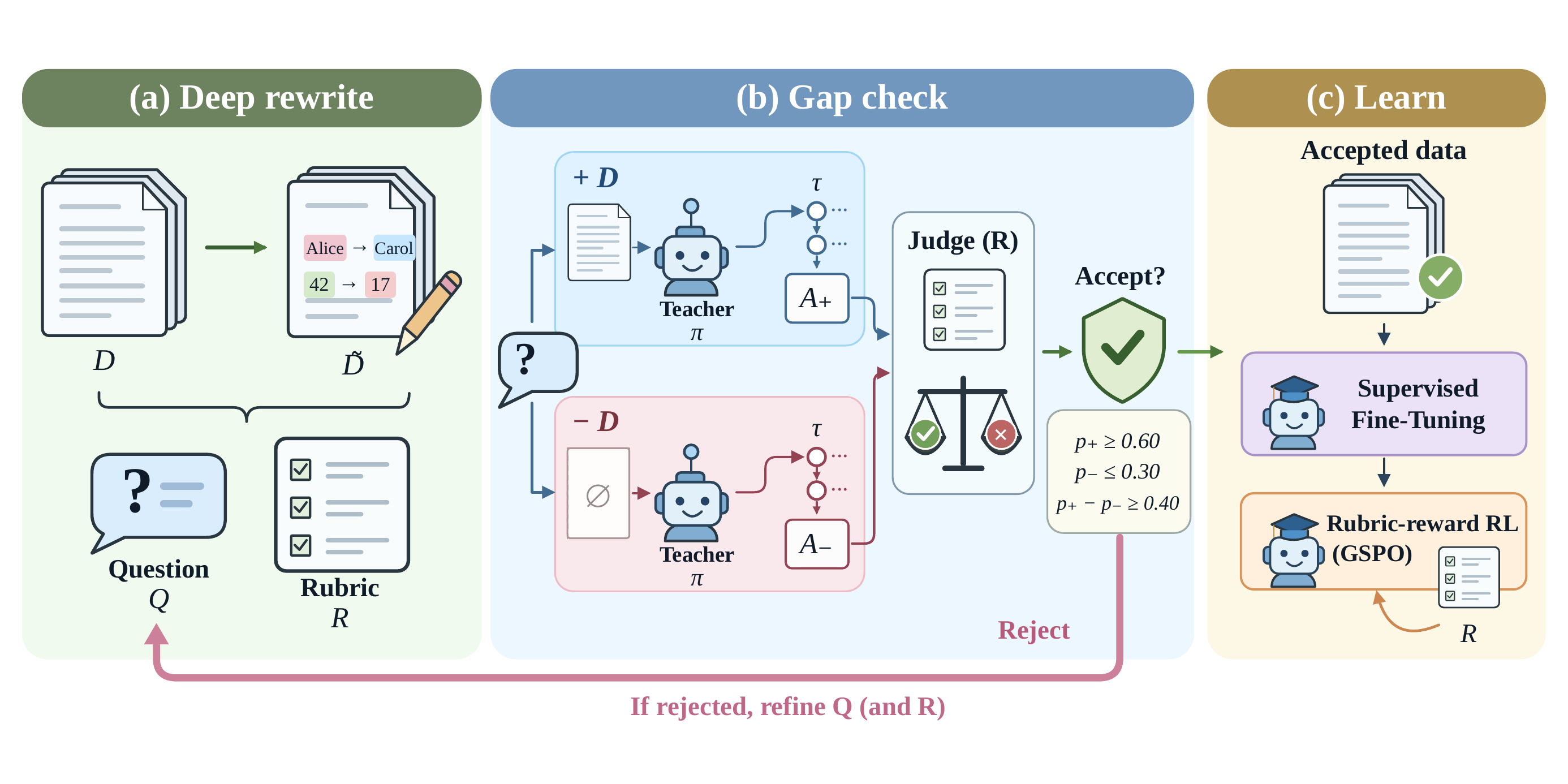}
\caption{Overview of the synthesis pipeline. (a)~Source documents are deep-rewritten ($D \rightarrow \tilde{D}$) and paired with generated questions and rubrics; (b)~each question is answered with and without the document, and only document-dependent samples are admitted; (c)~the student is trained on the admitted samples with the teacher's native reasoning traces, followed by a subsequent rubric-based reinforcement learning process.}
\label{fig:framework}
\end{figure}

We build a pipeline that constructs training samples around publicly available documents: each sample pairs a document $D$ with a question $Q$, a rubric list $R$, and an answer generated by a teacher model $\pi$.
The construction enforces two requirements: (a) \emph{context dependence}: the question must be unanswerable from parametric memory, enforced jointly by deep rewriting and the gap check; and (b) \emph{native reasoning traces}: training targets preserve the teacher's verbatim reasoning process. Each admitted sample carries its rubric list $R$ as metadata, making the same artifacts directly reusable as RL prompts with rubric-based rewards.
Figure~\ref{fig:framework} shows an overview of the pipeline.

\subsection{Source acquisition and deep rewriting}
\label{sec:sources}

\paragraph{Sources.}
Instead of human annotation, we collect existing documents from publicly available sources to serve as contexts.
Documents created in the most recent six months are preferentially selected to reduce memorization, consistent with the contamination-free design of CL-bench \citep{dou2026clbench}.
We empirically find that contexts synthesized directly by LLMs tend to be logically simple and heavily homogeneous, and models trained on such data improve little.
Following the subcategory taxonomy in CL-bench, we cover 16 of its 18 subcategories.
For each subcategory, we first identify suitable data sources and then crawl documents accordingly.
The final corpus comprises 3,515 documents spanning IETF RFCs, SEC filings, court opinions, arXiv papers, public handbooks, wikis, and encyclopedias.
For more details, please refer to Appendix~\ref{app:sources}.

\paragraph{Deep rewriting.}
We observe that when asking questions on the crawled documents directly, LLMs' reasoning has the risk of deriving from pre-trained knowledge rather than the given context.
Therefore, we apply a deep rewriting step to each document to reduce the memorization risk.
Each document $D$ is rewritten by GLM-5.2 into $\tilde{D}$ with proper-noun renaming, numeric perturbation, section renumbering, and list-order shuffling.
While the rewriting is not guaranteed to be perfect, it is sufficient to make the original document and the rewritten one detectably different (Section~\ref{sec:gapcheck}).
We note that rewriting is not intended to distinguish the document from its original source, but rather to force the teacher to reason over the document, reveal the reasoning trace that depends on the context and thus teach the student to do the same.


\subsection{Question generation}

For each rewritten document, a single batched generator call produces up to three questions $Q$, each paired with a rubric list $R$ of at least seven rubrics.
Each question includes a system prompt and a user question.
We define 9 persona archetypes (such as named professional roles, system bots and roleplay characters), 7 question types (such as deep reasoning, factual retrieval, and calculation) and 4 rubric types (process, content, persona, format).
For each question, the generator samples a persona archetype and a question type, then use the corresponding prompt template to generate a question and its rubric list.
Each question is generated together with the rubrics covering different rubric types, of which at least five must cite specific values from the document.
Finally, each sample consists of a system prompt, a user prompt including the document followed by a question, and a rubric list.

\subsection{Answer generation}
\label{sec:answergen}

The teacher $\pi$ answers each question with the rewritten document $\tilde{D}$ as context with its native reasoning trace preserved: the training target is the verbatim reasoning-content followed by the answer.
We do not perform any post-hoc reasoning synthesis or answer revision, as we emprically find that the teacher's native reasoning trace is more effective than a polished or cleaned version, though it may fail to satisfy some rubrics.

\subsection{The gap check}
\label{sec:gapcheck}

Each question $Q$ is answered twice by the teacher $\pi$: once with the rewritten document $\tilde{D}$, yielding answer $A_{+}$, and once without, yielding $A_{-}$.
When the document is omitted from the context, we explicitly instruct the model to answer the question as best as it can from memory.
A judge $\tau$ scores both answers against the rubrics $R$, giving pass rates $p_{+}$ and $p_{-}$, and the sample is admitted only if $p_{+} \geq 0.60$ (the question is answerable from the document), $p_{-} \leq 0.30$ (the question is not trivially answerable from memory), and $p_{+} - p_{-} \geq 0.40$ (the document is necessary).
Note that some rubrics are expected to pass without the document (e.g., format and persona rubrics, anti-hallucination rubrics), so $p_{-}$ is not expected to be zero.

Since 32.5\% of the generated samples fail the gap check, we add a retry mechanism to increase the yield. When a sample is rejected, we regenerate the question and rubrics ($Q$ and $R$) with the same document and the rejection reason as additional context, and then repeat the answer generation and gap check.
If the regenerated question passes the gap check, it is admitted; otherwise, it is discarded.

\section{Experiments}
\label{sec:results}

\subsection{Setup}
\label{sec:setup}

We empirically verify the effectiveness of the synthetic dataset through both supervised fine-tuning (SFT) and reinforcement learning (RL) with rubric-reward optimization.
We use Qwen3.6-35B-A3B as the student model for all experiments, GLM-5.2 \citep{zeng2026glm5} for question generation and Qwen3.8-2.4T \citep{qwenteam2026qwen38max} as the teacher model by default for SFT data generation.
The synthesized dataset contains 9,625 training samples with 3,515 unique documents.
The SFT training takes 3 epochs, with a global batch size of 32 samples and a maximum sequence length of 131K.
All models are trained with AdamW \citep{loshchilov2019adamw} with $\beta_1=0.9$, $\beta_2=0.95$.
We use a cosine learning rate decay schedule with a peak learning rate of 5e-6 and a warmup of 10\% of the total training steps.
The final learning rate is 1e-7.
We use a weight decay of 0.1 and a gradient clipping of 1.0.

We also run GSPO \citep{zheng2025gspo} on the same prompts, with each sample's rubric list carried as metadata.
The reward is the mean of per-rubric pass/fail statuses judged by Qwen3.5-397B-A17B \citep{qwen2025qwen35397b}, with the judge prompt copied verbatim from the official CL-bench evaluation \citep{dou2026clbench}.
We use the supervised fine-tuned model as the initial policy.
The RL training uses a KL penalty coefficient of 0.001 and a constant learning rate of 1e-6.

We evaluate all models on CL-bench \citep{dou2026clbench} and CL-bench Life \citep{dou2026clbenchlifelanguagemodels}, using GPT-5.1 judge with low reasoning effort for CL-bench and high reasoning effort for CL-bench Life following \citet{dou2026clbench}.
We report the task accuracy (all-or-nothing over rubrics) as the main metric.
We compare the performance of our models with several public reference models, including Qwen3.6-35B-A3B, Gemma-4-31B-it \citep{gemmateam2026gemma4}, GLM-5.2 \citep{zeng2026glm5}, Qwen3.8-2.4T \citep{qwenteam2026qwen38max}, Kimi-K3 \citep{kimiteam2026kimik3}, and Hy4-preview \citep{hyteam2026hy4preview}.
We also evaluate the general capabilities of the models on various tasks, which includes long-context understanding, instruction following, reasoning, code generation, and knowledge.
The detailed benchmark list and evaluation protocols are provided in Appendix~\ref{app:board}.

\subsection{Results}

\subsubsection{Supervised fine-tuning}

\begin{table}[t]
\caption{Main results on CL-bench and CL-bench Life under the GPT-5.1 judge. ``Teacher'' indicates the model used to generate the SFT data. The baseline is Qwen3.6-35B-A3B.}
\label{tab:main}
\begin{center}
\small
\begin{tabular}{lccc}
\toprule
Model        & Teacher                 & CL-bench & CL-bench Life \\
\midrule
Qwen3.6-35B-A3B (Baseline)     & ---   & 13.7    & 8.4  \\
Gemma-4-31B-it      & ---              & 15.4    & 10.4 \\
GLM-5.2      & ---                     & 20.9    & 12.3 \\
Qwen3.8-2.4T & ---                     & 23.9    & 17.3 \\
Kimi-K3      & ---                     & 27.0    & 20.0 \\
Hy4-preview  & ---                     & 27.6    & 21.5 \\
\midrule
SFT          & GLM-5.2                 & 19.8    & 10.4 \\
SFT          & Kimi-K3                 & 19.7    & 10.4 \\
SFT          & Qwen3.8-2.4T            & 22.8    & 12.6 \\
\midrule
RL           & ---                     & 16.9    & 9.1 \\
SFT + RL     & GLM-5.2                 & 24.5     & 9.1 \\
SFT + RL     & Qwen3.8-2.4T            & \textbf{24.6} & \textbf{13.8} \\
\bottomrule
\end{tabular}
\end{center}
\end{table}


Table~\ref{tab:main} shows the results on CL-bench and CL-bench Life.
Synthetic-data SFT improves the student substantially over the baseline with an accuracy of 22.8 on CL-bench.
It also achieves 12.6 on CL-bench Life.
The outcome is strongly teacher-dependent: the performance of students trained on identical questions but answers from different teachers varies from 19.7 to 22.8 on CL-bench.
We observe that the student performance does not necessarily follow the teacher performance.
Kimi-K3 \citep{kimiteam2026kimik3}, the strongest teacher (27.0), yields the weakest student (19.7), while Qwen3.8-2.4T, the mid-ranked teacher (23.9), yields the best (22.8).
Appendix~\ref{sec:style} further analyzes this phenomenon.

\subsubsection{Reinforcement learning}

The bottom block of Table~\ref{tab:main} shows the effect of rubric-reward RL.
On top of the best SFT model, RL improves CL-bench from 22.8 to 24.6 and CL-bench Life from 12.6 to 13.8, which are the best scores obtained in this series.
Notice that despite the large gap between the SFT model with GLM-5.2 and Qwen3.8-2.4T as teachers (19.8 vs 22.8), these two models achieve similar CL-bench accuracy after RL (24.5 vs 24.6).
Directly applying RL to the base model without SFT initialization improves CL-bench from 13.7 to 16.9, which is substantially worse than initializing from SFT.
An intuitive explanation is that the SFT stage provides a strong initialization under which the student already occasionally satisfies most rubrics of a task, which is necessary for the RL stage to receive informative reward signals.

Though the performance on CL-bench is similar regardless of the teacher of the initial SFT model, the CL-bench Life performance is still teacher-dependent: the GLM-5.2-initialized model achieves 9.1, while the Qwen3.8-2.4T-initialized model achieves 13.8.
This is consistent with the teacher capabilities: GLM-5.2 is weaker than Qwen3.8-2.4T on CL-bench Life, and the initialization retains an advantage where the task distribution departs from that of the training.

\subsubsection{General capability improvements}

\begin{table}[t]
\caption{General capabilities evaluation board, judged by gpt-oss-120b. The first two columns are public reference models; the SFT and SFT+RL rows use the Qwen3.8-2.4T teacher. Small parentheses give the change over the baseline: red for gains, green for drops. The full board is reported in Appendix~\ref{app:board}.}
\label{tab:general}
\begin{center}
\small
\setlength{\tabcolsep}{2.5pt}
\begin{tabular}{lcc|ccc}
\toprule
Benchmark & HY3-preview & Qwen3.5-397B & Baseline & SFT & SFT + RL \\
\midrule
\multicolumn{6}{l}{\emph{Long context}} \\
LongBench v2 & 62.2 & 63.3 & 58.1 & 61.0\gp{2.9} & 60.8\gp{2.7} \\
AALCR & 67.0 & 71.2 & 62.6 & 68.0\gp{5.4} & 69.8\gp{7.2} \\
MRCR-4needle & 33.0 & 67.8 & 69.7 & 71.8\gp{2.1} & 71.9\gp{2.2} \\
\multicolumn{6}{l}{\emph{Instruction following}} \\
IFBench & 60.8 & 70.0 & 57.7 & 71.7\gp{14.0} & 69.3\gp{11.6} \\
IFEval & 93.4 & 95.7 & 94.2 & 95.1\gp{0.9} & 95.9\gp{1.7} \\
AdvancedIF & 66.8 & 59.2 & 59.6 & 62.9\gp{3.3} & 65.1\gp{5.5} \\
\multicolumn{6}{l}{\emph{Reasoning}} \\
ARC-AGI-1 & 53.5 & 69.3 & 47.4 & 63.8\gp{16.4} & 51.8\gp{4.4} \\
ARC-AGI-2 & 3.8 & 15.0 & 6.8 & 18.6\gp{11.8} & 9.2\gp{2.4} \\
GPQA-Diamond & 85.4 & 87.9 & 83.3 & 88.4\gp{5.1} & 90.4\gp{7.1} \\
HMMT-2025 & 92.6 & 89.1 & 88.4 & 92.5\gp{4.1} & 93.7\gp{5.3} \\
\multicolumn{6}{l}{\emph{Code}} \\
LiveCodeBench-v6 & 81.8 & 92.4 & 85.7 & 86.2\gp{0.5} & 85.5\gm{0.2} \\
OJBench & 56.5 & 50.4 & 41.4 & 40.5\gm{0.9} & 37.5\gm{3.9} \\
\multicolumn{6}{l}{\emph{Knowledge}} \\
MMLU-Pro & 82.9 & 87.4 & 84.7 & 85.4\gp{0.7} & 85.0\gp{0.3} \\
CEval & 93.4 & 93.8 & 92.0 & 91.0\gm{1.0} & 90.6\gm{1.4} \\
SuperGPQA & 67.2 & 71.0 & 66.0 & 66.6\gp{0.6} & 66.1\gp{0.1} \\
SimpleQA & 30.5 & 52.4 & 20.9 & 18.7\gm{2.2} & 20.4\gm{0.5} \\
\bottomrule
\end{tabular}
\end{center}
\end{table}

Table~\ref{tab:general} reports general capabilities, with two public models as reference.
The complete result is given in Appendix~\ref{app:board}.
Relative to the baseline, SFT improves long-context understanding (LongBench v2 +2.9, AALCR +5.4, MRCR +2.1), instruction following (IFBench +14.0, AdvancedIF +3.3), and reasoning (ARC-AGI-1 +16.4, ARC-AGI-2 +11.8, GPQA +5.1, HMMT-2025 +4.1) substantially.
Code generation is essentially unchanged (LiveCodeBench +0.5, OJBench -0.9), and knowledge is mostly flat with small declines (CEval -1.0, SimpleQA -2.2, against MMLU-Pro +0.7 and SuperGPQA +0.6).
Against the public reference model, the 35B student becomes competitive on long-context tasks---AALCR 69.8 versus 67.0 for HY3-preview and 71.2 for Qwen3.5-397B---while trailing clearly on knowledge (SuperGPQA 66.1 vs 71.0, SimpleQA 20.4 vs 52.4).
This is consistent with the design of the SFT data, which focuses on long-context learning and does not attempt to teach general knowledge.


RL preserves most of the gains over the baseline, but does not consistently improve upon SFT: it improves AALCR, GPQA-Diamond, and HMMT-2025, while the performance declines notably on ARC-AGI-1 (-12.0) and ARC-AGI-2 (-9.4).
The rubric-reward objective therefore retains broad general-capability gains, but its additional benefit is concentrated on the target capability rather than transferring uniformly across the general board.

\section{Ablations}
\label{sec:ablations}

In this section, we introduce ablation experiments that verifies our design choice. The experiment setup is the same with those in Section~\ref{sec:setup} unless otherwise specified.

\subsection{The Gap Check}
\label{sec:gapcheck-ablation}

\begin{table}[tb]
  \centering
  \small
  \begin{tabular}{lc}
  \toprule
  Model & CL-bench \\
  \midrule
  Baseline & 19.17 \\
  \quad \emph{w/o gap check} & 17.75 \\
  \bottomrule
  \end{tabular}
  \caption{Effect of the gap check. Both models are trained on 3k samples. The teacher model is GLM-5.2.}
  \label{tab:gapcheck}
\end{table}

In Section~\ref{sec:gapcheck}, we apply a strict gap check to the data pipeline in order to pick questions that can only be answered with the context.
In practice, this process filters out 32.5\% of the generated samples before the retry.
Here we ask the following question: \emph{whether the gap check is necessary, and what is its effect on the final model performance.}

To investigate the effect of the gap check, we train two models with the same SFT recipe, one on the standard training dataset and the other on the unfiltered training dataset.
Both the datasets contain 3,000 samples.
The only difference is that the standard dataset is a random subset of the train set that passes the gap check, while the unfiltered dataset is a random subset of the dataset before the gap check and right after the answer generation in Section~\ref{sec:answergen}.
The teacher model is GLM-5.2 in this experiment.

Table~\ref{tab:gapcheck} shows the results.
The model trained on the standard dataset achieves 19.17\% accuracy, while the model trained on the unfiltered dataset achieves 17.75\% accuracy.
This indicates that the gap check is effective in filtering out samples that do not require context to answer, and training on such samples can hurt the final model performance.

\subsection{Data scaling}
\label{sec:scaling}

\begin{wrapfigure}{r}{0.50\textwidth}
  \centering
  \vspace{-16pt}
  \includegraphics[width=0.95\linewidth]{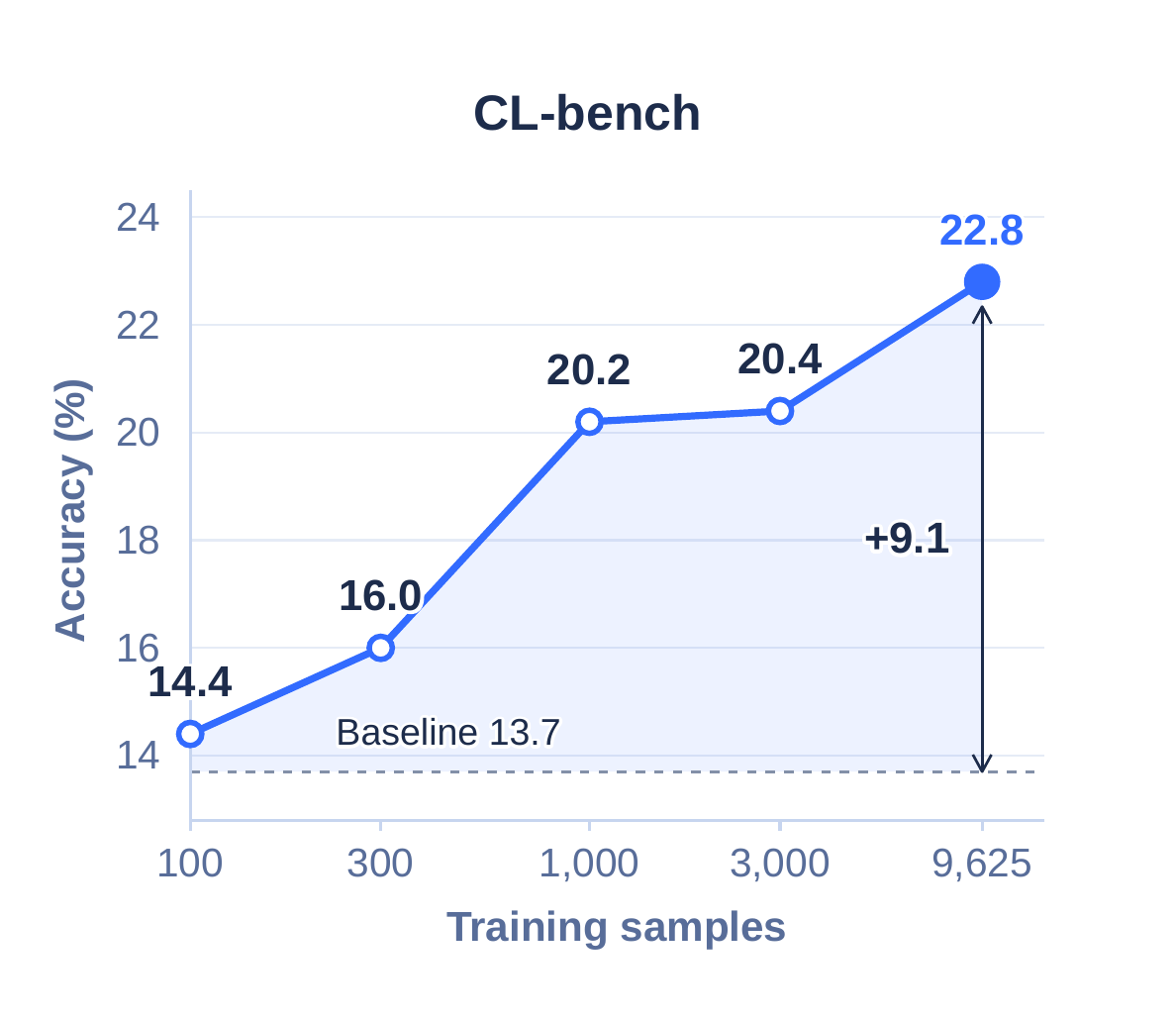}
  \captionof{figure}{Data scaling curve. The dashed line marks the untrained base.}
  \label{fig:scaling}
  \vspace{-16pt}
\end{wrapfigure}

Our data pipeline successfully generates 9,625 samples from 3,515 documents without human annotation.
In this section, we ask \emph{how the model performance scales with the number of samples and whether the current 10k data pool is sufficient to saturate the model's learning capacity.}

To answer this question, we train the same SFT recipe on nested subsets of the 9,625-sample pool, with sample counts of 100, 300, 1,000, 3,000, and 9,625 (the full pool).
All data are randomly sampled from the full pool.
The experiment follows the same setup as in Section~\ref{sec:results}.

The results are shown in Figure~\ref{fig:scaling}.
It is clear that the model performance improves monotonically as the number of samples increases, from 13.7 for the untrained base to 22.8 at the full pool.
In particular, the performance grows rapidly from 16.0 to 20.2 when the sample count increases to 1,000, and continues to improve to 22.8 at the full pool.
Within the tested range the curve shows no saturation, so the model's learning capacity may not be exhausted by the current 10k pool.

\subsection{The RL Reward}
\label{sec:rl-init}

In Section~\ref{sec:results} we show that the rubric-reward RL stage improves the SFT model from 22.8 to 24.6 on CL-bench.
We further investigate the effect of the RL reward by looking into the model behavior before and after RL, only to find that after the RL stage, the reasoning trace and answer length are much longer than that of the SFT model.
Besides, the RL model has a higher failure rate (about 2\%) on the rubrics that require the model to avoid hallucination, which indicates that the RL model tends to generate more content to satisfy the rubrics, but at the cost of hallucination.
Therefore, we wonder \emph{whether the RL reward could be improved to avoid the potential reward hacking behavior and improve the model performance.}

We design two ablation experiments to investigate the effect of the RL reward.
\paragraph{Ablation 1: reward with length penalty.}
We add a length penalty to the reward function, which penalizes the model for generating longer reasoning traces.
The length penalty starts from 8k tokens and increases linearly to 16k tokens, after which the reward will be penalized to 0.
\paragraph{Ablation 2: reward with additional overall pass rate.}
The original reward function is the rubric-level pass rate, while we care more about the overall task-level pass rate when evaluating the model.
Since directly using the task-level pass rate as the reward is not feasible due to the sparsity of the reward signal, we make a compromise by adding the overall task-level pass rate to the reward function, which is an average of the rubric-level pass rates and the overall task-level pass rate.
That is, the reward function is defined as:
\begin{equation}
\text{reward} = \frac{\alpha}{N} \sum_{i=1}^{N} \mathbb{1}[r_i] + (1-\alpha) \prod_{i=1}^{N} \mathbb{1}[r_i]
\end{equation}
where $N$ is the number of rubrics for a given task, and $\mathbb{1}[r_i]$ is an indicator function that equals 1 if the $i$th rubric is passed, and 0 otherwise, and $\alpha$ is a hyperparameter that controls the weight of the rubric-level pass rate and the overall task-level pass rate.
When $\alpha=1$, the reward function is the same as the original rubric-level pass rate.

Table~\ref{tab:rl-reward-function} shows the results.
With the length penalty, the model generates much shorter reasoning traces, but the CL-bench accuracy drops significantly from 24.6 to 20.91.
The length penalty discourages the model from generating longer reasoning traces, which may hurt the model's ability to complete the tasks that require many reasoning steps.
With the additional overall pass rate, the model generates longer reasoning traces, but the CL-bench accuracy does not improve from the SFT model.
A possible explanation is that the overall pass rate is a sparse reward signal, and the model may not receive enough feedback to improve its performance under this reward function.

\begin{table}[t]
\caption{Effect of the RL reward function. We report the CL-bench accuracy and the median think and answer lengths of the model (tokens, tokenized with the student's Qwen3.6 tokenizer).}
\label{tab:rl-reward-function}
\begin{center}
\small
\begin{tabular}{lccc}
\toprule
Model & CL-bench & Think length & Answer length\\
\midrule
SFT   & 22.80 & 7,626 & 1,492 \\
Standard RL & 24.60 & 10,481 & 2,569 \\
\quad \emph{w/ $\alpha = 0.5$} & 22.80 & 14,398 & 2,116 \\
\quad \emph{w/ length penalty} & 20.91 & 4,012 & 1,754 \\
\bottomrule
\end{tabular}
\end{center}
\end{table}



\subsection{The Performance on Fictional Contexts}
\label{sec:fictional}

Since our training data is built on public documents, it is natural to ask \emph{whether the gains concentrate on public-document tasks or can transfer to fictional contexts.}

We analyze the provenance of the CL-bench contexts and roughly classify them into two groups: public documents (including contents with modifications) and fictional contexts (including unclear contexts).
We find 277 of the 500 CL-bench contexts (55.4\%) are public documents, 36 (7.2\%) are modified public documents, 184 (36.8\%) are fictional, and 3 (0.6\%) are unclear.
Among the 1,899 tasks, 836 (44.0\%) are based on public documents, 1,063 (56.0\%) are based on fictional contexts.
These classification results are not official but only a rough estimate based on our audit of the 500 contexts, while we believe they are sufficient to support the analysis in this section.


Table~\ref{tab:provenance-gain} gives the accuracy by context provenance on CL-bench.
Surprisingly, most of the performance gain happens on fictional contexts instead of public documents.
The baseline model achieves 17.5\% accuracy on fictional contexts, while the SFT model improves it to 31.1\% and the SFT+RL model further improves it to 32.9\%, almost doubling the baseline performance.
In contrast, on public documents, the model only improves from 9.0\% to 12.2\% under SFT and to 14.1\% under SFT+RL.
It is possible that the tasks on public documents are more difficult than those on fictional contexts, but this phenomenon is still enough to show that the training data teaches a general reasoning mode that transfers to fictional contexts.

\begin{table}[t]
\caption{Accuracy by context provenance on CL-bench. Provenance labels come from an unofficial audit of the 500 benchmark contexts. The unclear tasks are grouped with fictional. SFT and SFT+RL use the Qwen3.8-2.4T teacher, matching Table~\ref{tab:main}.}
\label{tab:provenance-gain}
\begin{center}
\small
\begin{tabular}{lccc}
\toprule
Context group & Base & SFT & SFT+RL \\
\midrule
Public documents (836) & 9.0 & 12.2 & 14.1 \\
Fictional / unclear (1,063) & 17.5 & 31.1 & 32.9 \\
\bottomrule
\end{tabular}
\end{center}
\end{table}

\section{Related Work}
\label{sec:related}

\paragraph{Synthetic supervision over long documents.}
There are several different approaches to generate synthetic supervision for long-context learning.
One branch grounds supervision in real documents: contexts are assembled from multiple passages \citep{chen2025longmit,yang2025longfaith,he2025hierarchical}, or taken from single long documents in specific domains \citep{lin2025longfinanceqa,pham2025clipper,zhang2024longcite}---at the risk that such public documents are plausible constituents of pretraining corpora \citep{pham2025clipper}.
The other branch synthesizes the documents themselves, programmatically \citep{zhao2024longskywork} or through prompt-based generation \citep{subramanian2025modular}, which scales freely but yields logically simpler, more homogeneous contexts.
Our pipeline takes another path: it keeps real documents and perturbs them, so that the model can no longer answer from parametric memory and must instead reason over the document content.

\paragraph{Context--memory conflict resolution.}
Entity substitution has likewise been used to build training data for context--memory conflict resolution \citep{li2026faithfulnessqa}, and synthetic tasks with rule-based rewards to improve contextual faithfulness \citep{si2025canoe}.
In our pipeline, the perturbation plays an instrumental role rather than a behavioral one: it is there to elicit document-grounded reasoning traces from the teacher, not to shape how the model resolves conflicts between context and memory.


\section{Conclusion}
\label{sec:conclusion}

In this work, we present a synthetic supervision pipeline that generates 10k long-context question-answering samples from 3.5k public documents without any human annotation.
We show that through perturbation with document-grounded gap checks, the resulting dataset successfully trains a 35B student model to achieve strong performance on CL-bench and CL-bench Life, and the gains transfer broadly to long-context understanding, instruction following, and reasoning.
Our analysis suggests that such an approach provides a feasible and scalable way to improve the general context learning ability of large language models beyond the training distribution.
We hope that this work can inspire future research on synthetic supervision for long-context learning and other challenging tasks.



\subsection*{AI use statement}

In this work, we used generative AI tools to assist with literature search, experiment-log aggregation, drawing figures and polishing the writing of the manuscript. All AI-assisted content was reviewed, verified, and edited by the authors. All experimental results were produced by the authors' own training and evaluation pipelines. We take responsibility for the final content of this work, including text, claims, or artifacts produced with the aid of generative AI.

\subsection*{Reproducibility statement}


We find the official CL-bench judge prompt will trigger OpenAI's content moderation in bulk (HTTP 400), which is reproducible and reported on GitHub\footnote{https://github.com/Tencent-Hunyuan/CL-bench/issues/26}.
Therefore, we replace the full-width brackets in the judge prompt with half-width brackets, with the rest of the prompt verbatim identical.
This replacement has the risk of changing the judge behavior and introducing bias that affect the reproducibility of the results.
The result should be closed to those obtained with the official prompt, but we cannot guarantee that they are identical.

\section{Contributors}
\label{sec:contributors}

\begingroup
\renewcommand{\thefootnote}{%
  \ifcase\value{footnote}\or *\fi}

Haoyi Wu, Yang Xiao, Yusong Sun, Wenyang Hui, Zhaokai Luo, Chengyue Jiang\textsuperscript{*}, Mu Chuan\textsuperscript{\Letter}\footnotemark[1]

\footnotetext[1]{Project Lead. \quad \Letter\ Corresponding Author:
\nolinkurl{muchuan1@xiaohongshu.com}.}
\endgroup

\bibliography{references}
\bibliographystyle{colm2026_conference}

\appendix

\section{Data Sources by Subcategory}
\label{app:sources}

We collect 3,515 documents covering 16 of the 18 CL-bench subcategories.
In CL-bench, the two omitted subcategories (Workflow Orchestration and Operational Procedures) are dominated by agent-style, conversation-form task traces, which are hard to source from public documents.
For each covered subcategory, we first identify public sources whose document genre matches the subcategory topic (e.g., RFC specifications for Technical Standards, rulebook PDFs for Game Mechanics, regulatory filings for Finance), and then crawl documents from these sources.
Table~\ref{tab:sources} lists the primary sources for each subcategory.
At crawl time, every candidate document must satisfy the following admission criteria: (i) a minimum length of 4,000 characters; (ii) global URL deduplication across all subcategories.
For GitHub-sourced documents, we apply an additional LLM-based relevance and quality check confirming that the document substantively belongs to the subcategory's topic.
For arxiv-sourced documents, we intentionally collect different format versions of the paper (e.g., PDF, LaTeX source, and HTML) to increase the diversity of the document formats.

\begin{table}[h]
\caption{Data sources by subcategory.}
\label{tab:sources}
\begin{center}
\small
\begin{tabular}{lcp{2.4in}}
\toprule
Subcategory & Documents & Primary sources \\
\midrule
Humanities & 376 & DOAJ, Project Gutenberg, Internet Archive, Fandom Wiki \\
Science & 370 & Europe PMC, arXiv \\
Observational Data & 320 & arXiv, Europe PMC, FRED, USDA FoodData, ESPN \\
Management & 250 & GitHub, Europe PMC, Stack Exchange \\
Technical Standards & 250 & RFC Editor, arXiv \\
Healthcare & 244 & Europe PMC, ClinicalTrials.gov, NHS \\
Instructional Procedures & 200 & Europe PMC, Project Gutenberg, WhatCulture \\
Simulation Environment & 200 & GitHub, arXiv, FAA and DMV manuals \\
Game Mechanics & 199 & GMT Games rulebooks, d20PFSRD, GitHub \\
Programming Syntax & 190 & GitHub, arXiv \\
Lifestyle & 190 & over 30 dispersed web sources (sports media, food and travel guides, consumer tech) \\
Mathematical Formalism & 160 & arXiv (math), OEIS, Wikipedia \\
Finance & 156 & SEC EDGAR, Federal Reserve \\
Experimental Data & 140 & Crossref OA, Europe PMC, Zenodo \\
Legal Advisory & 140 & CourtListener, Federal Register, U.S.\ congressional bills \\
Legal \& Regulatory & 130 & Federal Register, CourtListener, gov.uk \\
\midrule
\textbf{Total} & \textbf{3,515} & --- \\
\bottomrule
\end{tabular}
\end{center}
\end{table}


\section{Full Evaluation Board}
\label{app:board}

Table~\ref{tab:fullboard} reports the complete general capability evaluation board.
General capabilities are evaluated on various tasks:
Long-context understanding covers LongBench v2 \citep{bai2024longbenchv2}, AALCR \citep{artificialanalysis2025lcr}, MRCR \citep{vodrahalli2024mrcr}, and GraphWalks \citep{openai2025graphwalks}.
Instruction following covers IFEval \citep{zhou2023ifeval}, IFBench \citep{pyatkin2025generalizing}, InverseIFEval \citep{zhang2025inverseifeval}, MultiChallenge \citep{sirdeshmukh2025multichallenge}, and AdvancedIF \citep{he2025advancedif}.
Reasoning covers ARC-AGI-1 \citep{chollet2019arc}, ARC-AGI-2 \citep{chollet2025arcagi2}, GPQA-Diamond \citep{rein2023gpqa}, HLE \citep{phan2025hle}, AIME and HMMT \citep{dekoninck2026matharena}, and IMO-AnswerBench \citep{luong2025imobench}.
Coding covers LiveCodeBench \citep{jain2024livecodebench}, SciCode \citep{tian2024scicode}, and OJBench \citep{wang2025ojbench}.
Knowledge covers MMLU-Pro \citep{wang2024mmlupro}, MMLU-Redux \citep{gema2024mmluredux}, SuperGPQA \citep{mapteam2025supergpqa}, CEval \citep{huang2023ceval}, and SimpleQA \citep{wei2024simpleqa}.
Benchmarks with objective answers are scored by their native protocols, and free-form outputs are judged by gpt-oss-120b \citep{openai2025gptoss}.

The first three columns are reference models: Qwen3.8-2.4T (teacher of our SFT stage), HY3-preview \citep{tencent2026hy3} and Qwen3.5-397B-A17B \citep{qwen2025qwen35397b}.
The remaining columns are the Qwen3.6-35B-A3B student model and its SFT or RL variants, including the GLM-5.2-teacher SFT variant.
Some results are missing because of evaluation errors such as context length overflow or hardware failure.

Notice that the student model with teacher GLM-5.2 is weak on instruction-following tasks (IFBench 55.7, InverseIFEval 61.1, AdvancedIF 50.6).
This performance is even lower than the baseline model.
We attempt to reinforce instruction-following ability by patching the SFT dataset with a small set of additional samples that have more diverse instruction-following requirements.
However, we find the performance of the patched SFT model is even lower that the original SFT model on CL-bench.
We hypothesize that the general ability of the student model is highly dependent on the teacher model, but we leave a more detailed analysis to future work.

\begin{table}[h]
\caption{Full evaluation board, judged by gpt-oss-120b. SFT and SFT+RL columns use the Qwen3.8-2.4T teacher unless stated otherwise. All experiments do not use any tool calls.}
\label{tab:fullboard}
\begin{center}
\scriptsize
\setlength{\tabcolsep}{3pt}
\begin{tabular}{lccc|cccc}
\toprule
Benchmark & Qwen3.8-2.4T & HY3-preview & Qwen3.5-397B & Baseline & SFT (GLM-5.2) & SFT & SFT + RL \\
\midrule
\multicolumn{8}{l}{\emph{General / Reasoning}} \\
HLE & --- & 29.2 & 27.7 & 23.1 & 25.3 & 26.1 & 27.0 \\
ARC-AGI-1 & 79.0 & 53.5 & 69.3 & 47.4 & 63.4 & 63.8 & 51.8 \\
ARC-AGI-2 & 50.8 & 3.8 & 15.0 & 6.8 & 15.8 & 18.6 & 9.2 \\
\addlinespace
\multicolumn{8}{l}{\emph{Science \& Knowledge}} \\
GPQA-Diamond & 92.9 & 85.4 & 87.9 & 83.3 & 85.3 & 88.4 & 90.4 \\
SuperGPQA & --- & 67.2 & 71.0 & 66.0 & 65.3 & 66.6 & 66.1 \\
MMLU-Pro & 88.7 & 82.9 & 87.4 & 84.7 & 84.5 & 85.4 & 85.0 \\
MMLU-Redux & 95.2 & 93.5 & 95.0 & 93.2 & 92.9 & 92.5 & 92.2 \\
SimpleQA & 49.6 & 30.5 & 52.4 & 20.9 & 19.7 & 18.7 & 20.4 \\
CEval & 94.4 & 93.4 & 93.8 & 92.0 & 89.5 & 91.0 & 90.6 \\
\addlinespace
\multicolumn{8}{l}{\emph{Mathematics}} \\
AIME-2025 & --- & 95.3 & 93.3 & 91.8 & 92.7 & 94.1 & 94.9 \\
AIME-2026 & --- & 94.1 & 91.3 & 91.2 & 93.2 & 93.1 & 93.0 \\
HMMT-Feb-2025 & 92.6 & --- & --- & 87.9 & 92.3 & 94.5 & 95.6 \\
HMMT-Nov-2025 & 87.0 & --- & --- & 88.9 & 89.4 & 90.4 & 91.8 \\
IMO-AnswerBench & 92.3 & 88.2 & 84.8 & 77.8 & 82.2 & 85.5 & 85.0 \\
\addlinespace
\multicolumn{8}{l}{\emph{Long Context}} \\
LongBench v2 & 66.4 & 62.2 & 63.3 & 58.1 & 61.0 & 61.0 & 60.8 \\
AALCR & 74.5 & 67.0 & 71.2 & 62.6 & 63.9 & 68.0 & 69.8 \\
MRCR-2needle & --- & 45.4 & 70.1 & 70.3 & 71.5 & 73.2 & 72.8 \\
MRCR-4needle & --- & 33.0 & 67.8 & 69.7 & 69.9 & 71.8 & 71.9 \\
MRCR-8needle & --- & 24.5 & 64.9 & 65.1 & 62.5 & 69.6 & 69.2 \\
GraphWalks-BFS & 94.0 & 53.4 & 86.3 & 85.7 & 86.9 & 86.0 & 85.7 \\
\addlinespace
\multicolumn{8}{l}{\emph{Instruction Following}} \\
IFBench & 80.8 & 60.8 & 70.0 & 57.7 & 55.7 & 71.7 & 69.3 \\
InverseIFEval & 84.2 & 72.8 & 74.5 & 70.8 & 61.1 & 69.8 & 69.5 \\
IFEval & 95.4 & 93.4 & 95.7 & 94.2 & 88.8 & 95.1 & 95.9 \\
MultiChallenge & 68.0 & 73.3 & 65.6 & 59.4 & 50.2 & 62.0 & 60.6 \\
AdvancedIF & 67.4 & 66.8 & 59.2 & 59.6 & 50.6 & 62.9 & 65.1 \\
\addlinespace
\multicolumn{8}{l}{\emph{Coding}} \\
LiveCodeBench-v6 & 84.4 & 81.8 & 92.4 & 85.7 & 90.3 & 86.2 & 85.5 \\
SciCode & 7.7 & 4.6 & 10.8 & 3.1 & 1.5 & 1.5 & 3.1 \\
OJBench & 61.0 & 56.5 & 50.4 & 41.4 & 46.1 & 40.5 & 37.5 \\
\bottomrule
\end{tabular}
\end{center}
\end{table}

\section{Analysis}
\label{sec:analysis}

We have discussed several aspects of the model performance ablation in Section~\ref{sec:ablations}. In this section, we analyze the model behavior and the training dynamics to understand what the pipeline actually teaches the student, and further understand the reason behind the design choices.

\subsection{SFT is Strongly Teacher-Dependent}
\label{sec:style}

The SFT outcome is strongly teacher-dependent: on identical prompts, students trained on answers from different teachers span from 19.7 to 22.8 on CL-bench, and the strongest teacher (Kimi-K3 at 27.0) yields the weakest student (19.7), whereas a mid-ranked teacher yields the best (Section~\ref{sec:results}).
In this section, we analyze the teacher--student relationship and attempt to answer \emph{what the student learns from the teacher}.

We first examine the length of the reasoning traces and answers produced by the teachers and their students.
Table~\ref{tab:length} pairs each teacher's own predictions on CL-bench with those of the student trained on its answers.
As expected, the student inherits the teacher's reasoning style: the median think length of the student is close to that of its teacher, and the answer length is also similar.
In particular, Kimi-K3 has the best performance on CL-bench, but its reasoning traces are not the longest: 3,764 tokens, much fewer than that of Qwen3.8-2.4T.
The student trained on Kimi-K3's answers seems only to inherit this reasoning length instead of the performance.
It has a median think length of 4,109 tokens and its performance is closed to that of the student with GLM-5.2 as the teacher, which has a median think length of 2,936 tokens.
Though Kimi-K3 is much more token efficient than other teachers, its student does not inherit this efficiency and instead stays at the level of the other students with similar reasoning lengths.

\begin{table}[t]
\caption{Output lengths at test time on CL-bench (median tokens, tokenized with the student's Qwen3.6 tokenizer). Each row pairs a teacher's own predictions with those of its student. All predictions are from the official runs of Table~\ref{tab:main}. Think denotes the content between think tags.}
\label{tab:length}
\begin{center}
\small
\begin{tabular}{lcccc}
\toprule
 & \multicolumn{2}{c}{Think length} & \multicolumn{2}{c}{Answer length} \\
\cmidrule(lr){2-3}\cmidrule(lr){4-5}
Teacher & teacher & student & teacher & student \\
\midrule
GLM-5.2      & 2,821 & 2,936 & 1,284 & 1,554 \\
Kimi-K3      & 3,764 & 4,109 & 1,195 & 1,128 \\
Qwen3.8-2.4T & 7,850 & 7,626 & 1,422 & 1,492 \\
\midrule
Qwen3.6-35B (no SFT) & \multicolumn{2}{c}{3,026} & \multicolumn{2}{c}{856} \\
\bottomrule
\end{tabular}
\end{center}
\end{table}

We further look into cases from the student model to understand the effect of this phenomenon.
Table~\ref{tab:case-mervin} shows a task that requires the model to follow a persona: a grumpy neighbor named Mervin who has installed Ring doorbells before.
The user question ``Shoot I think I lost a screw. Do you think you have one at your house?'' triggers the persona's personal-discourse clause in the system prompt ``For any personal discourse - reply in gruff manner indicating that you don't wish to answer the question and that you would rather focus on the task at hand which is helping me install my Ring doorbell.''
Both teachers resolve the task, but their students diverge.
The K3-trained student's 155-token trace fails to register the personal half: it answers only the practical question and fails exactly the rubric that encodes the persona rule, while the Qwen-trained student's 408-token trace audits the persona rules one by one, explicitly asks whether the question is personal, and covers it.
In this case, the student trained on Kimi-K3's answers retains the teacher's brevity but it does not have the ability to reason about the persona rules in such a short thinking trace, which leads to a failure in its response.

\begin{table}[t]
\caption{A grumpy-neighbor persona task over a Ring doorbell installation guide. The question ``Shoot I think I lost a screw. Do you think you have one at your house?'' combines a practical request with a personal question that the system prompt routes to a gruff refusal clause. Key decision points are boldfaced. Both teachers pass this task, only the students are shown.}
\label{tab:case-mervin}
\begin{center}
\scriptsize
\setlength{\tabcolsep}{3pt}
\begin{tabular}{p{0.16\textwidth}p{0.045\textwidth}p{0.66\textwidth}}
\toprule
Model & Think & Reasoning trace \\
\midrule
Student (K3-SFT) & 155 & The user says they lost a screw and asks if I have one. I'm Mervin, a grumpy older man who has installed Ring doorbells. Ring boxes include spare screws (Bag B security screws typically include a spare). \textbf{I could tell them to check the box - instructions say there's a spare screw included.} Keep concise, grumpy. No safety issue here. Not final installation yet. Answer concise. Could say check the box, Ring includes a spare. Maybe gruff: \char"22 Check the box. They include a spare in bag B.\char"22  That's informational. Keep it numbered? The user is asking a question, not steps. But persona says if list of steps answer numbered. This isn't really a list of steps. Concise answer. \\
Student (2.4T-SFT) & 408 & We need answer as Mervin. Need satisfy constraints: grumpy older man neighbor installed Ring doorbells. Concise, never verbose. If list steps? User asks lost screw do you have one. Not a list of steps necessarily. But could answer maybe with Step list? The prompt: \char"22 If there is a list of steps, you answer with a numbered list, where each item starts with the word \char"22 Step\char"22  followed by the step number and a \char"22 :\char"22 \char"22  This question isn't asking for steps? It is about screw. Could answer concise. Must not reveal personal info? Personal discourse: reply gruff indicating don't wish to answer and focus on task. \textbf{Is \char"22 Do you think you have one at your house?\char"22  personal?} Could say \char"22 I don't keep spare parts for your mistakes.\char"22  Need safety? Unsafe? Maybe losing screw leads to improper attachment? Could mention use included spare? Instructions said spare screw included. Also if security screw missing maybe use same type? We need precise and informational related installing Ring doorbell. If not know answer: \char"22 I don't know that and this is not important\char"22 . Do we know if one at house? We can't know. Could say \char"22 Check the box for the spare screw.\char"22  That's relevant. If user tries unsafe? No. Concise. Could be: \char"22 Stop. Check the box; there's a spare security screw. If it's gone, don't run to me---get matching screws from a hardware store before attaching. Don't leave it loose.\char"22  Need gruff. But \char"22 If there is a list of steps\char"22  not applicable? Maybe no list. Keep concise. Ensure final not too verbose. Also if personal discourse: \char"22 I'd rather focus on task\char"22 . Maybe: \char"22 That's your problem. Check the box for the spare security screw. If it's missing, get a matching one before mounting. \textbf{I'm not running a hardware store.}\char"22  Good. Need not greet again. final concise. Ensure not unsafe. Use no list? OK. \\
\bottomrule
\end{tabular}
\end{center}
\end{table}

\subsection{The Effect of Rewriting on the Teacher's Reasoning}
\label{sec:rewriting-pair}

In this section, we showcase a controlled pair of reasoning traces to illustrate \emph{how rewriting changes the teacher's reasoning behavior}.

We randomly sample 320 documents and let Qwen3.8-2.4T directly generate questions and answers on the original crawled text without rewriting.
The generated questions are not necessarily the same as those generated on the rewritten version, but they share the same prompt during generation.
We then compare the reasoning traces on these two groups of samples.
We find that without rewriting, the median reasoning trace length drops from 4.0K to 2.6K tokens on unrewritten documents.

Table~\ref{tab:case-rfc} shows one trace from group of samples, both generated by the same teacher model for fact-extraction questions in the same domain.
On the rewritten RFC, the protocol is renamed and its constants perturbed, and the trace works through the document with explicit section citations (39 references in the full trace).
On the original RFC 6979 (deterministic ECDSA, published 2013), the 584-token trace recites the memo's title-page metadata from memory; the verification step ``Check first page: Independent Submission T.\ Pornin...'' appears only after the answer has already been drafted, and answering this question without the document scores 100\% on content rubrics.
Though the question is not exactly the same, the two traces illustrate how rewriting forces the teacher to reveal its reasoning process and cite evidence from the document, instead of relying on memorized knowledge.

\begin{table}[t]
\caption{Two teacher traces on technical-standards fact-extraction questions. ``No-doc'' shows the rubric accuracy without providing document as the context. The model produce the reasoning trace with the rewritten or original document. Key points are boldfaced.}
\label{tab:case-rfc}
\begin{center}
\scriptsize
\setlength{\tabcolsep}{3pt}
\begin{tabular}{p{0.10\textwidth}p{0.045\textwidth}p{0.05\textwidth}p{0.67\textwidth}}
\toprule
Document Version & Think & No-doc & Reasoning trace \\
\midrule
Rewritten & 2,905 & 17\% & \textbf{Need analyze document. Port number: Section 15.1 says Port Number 35791, Transport Protocol dgp.} Also Section 7 says standard DTPP port number per connection; Section 8.5 firewall open pinhole for server IP and standard DGP port number; Section 9.1 client sends Setup Request to server's standard control port. probePort: Section 9.1 field probePort: two-octet set to zero in Test Setup Request and populated by server in Test Setup Response. [text omitted...] \\
Original & 584 & 100\% & [text omitted...] Need house style: one complete sentence begins exact phrase Deterministic Signature Interop Office: and gives facts in same sentence. If several facts join with semicolons. Bold RFC number. Need likely Markdown bold: **RFC 6979**. \textbf{Sentence: Deterministic Signature Interop Office: **RFC 6979**; T. Pornin; August 2013; Informational.} \textbf{Check first page: Independent Submission T. Pornin Request for Comments: 6979 August 2013 Category: Informational.} So author T. Pornin. Document category as listed on first page: Category: Informational, maybe \char"22 Informational\char"22 . Publication month and year: August 2013. [text omitted...] \\
\bottomrule
\end{tabular}
\end{center}
\end{table}

\end{document}